\documentclass{article}

\usepackage{arxiv}

\usepackage[T1]{fontenc} 
\usepackage{hyperref}     
\usepackage{amsfonts}     
\usepackage{nicefrac}      
\usepackage{microtype}     
\usepackage{lipsum}
\usepackage{float}

\usepackage{tabularx,booktabs,ragged2e}
\usepackage{rotating}                      
\usepackage{newtxmath}                     
\usepackage[inline]{enumitem}              
\usepackage{caption,subcaption}            
\usepackage{eurosym}                       
\usepackage{xcolor}                        
\usepackage{amsmath}

\begin{document}

\title{Expected Possession Value of Control and Duel Actions for Soccer Player's Skills Estimation}

\author{Andrei Shelopugin \\
Independent Researcher \\
shelopuginandrey@gmail.com}

\maketitle

\begin{abstract}

Estimation of football players' skills is one of the key tasks in sports analytics. This paper introduces multiple extensions to a widely used model, expected possession value (EPV), to address some key challenges such as selection problem. First, we assign greater weights to events occurring immediately prior to the shot rather than those preceding them (decay effect). Second, our model incorporates possession risk more accurately by considering the decay effect and effective playing time. Third, we integrate the assessment of individual player ability to win aerial and ground duels. Using the extended EPV model, we predict this metric for various football players for the upcoming season, particularly taking into account the strength of their opponents.

\end{abstract}


\section{Introduction}

Estimating players' skills is a key challenge for football managers, scouts, and analysts. While video analysis can offer insights \cite{bergkamp2022soccer}, conducting it for all players in the market is impractical. Analysts must therefore use filtering mechanisms to narrow down the selection pool. This can involve basic profile data like age, position, and nationality. Alternatively, leveraging statistics or metrics correlated with actual player performance offers a more effective approach. By using these metrics, analysts can make more informed decisions in player selection.

Currently, accurately assessing the skill levels of players based on historical data has become a complex task. Soccer, being a complicated game, presents various challenges in this regard, one of which is the fact that out of the 22 players on the field, only one possesses the ball at any given time. In recent years, sports analytics, including soccer analytics, has made significant advancements in the understanding how to effectively estimate players' abilities, particularly in terms of ball possession skills.

In the past, player analysis predominantly relied on basic statistics such as goals scored, successful passes, and fouls committed. However, there has been a shift towards utilizing advanced metrics, which provide a more comprehensive and nuanced evaluation of player performance.

The challenge with these metrics lies in their reliance on historical data, limiting their use in predicting future performance. While they may work well in "closed" leagues like the NBA, where teams share similar strengths, soccer's diverse leagues present a more complex scenario. This complicates player selection, as managers must understand how players will adapt to new clubs or leagues. Analysts need to assess the resistance levels of different leagues and the stylistic characteristics of new clubs for optimal player hiring decisions.

In this paper, we modify the expected possession value model. Using this model, we calculate a reward, accumulated by each player during the season, that characterizes a footballer's ability to play with the ball and predict the values of this reward for the next season.

\section{Related Work}

J. Hollinger introduced the Player Efficiency Rating ($PER$)\cite{hollinger2005pro} as a metric for evaluating NBA players, utilizing basic statistics. He introduced the concept of the value of possession and suggested a formula that rewards and punishes players for successful and unsuccessful actions.

Pollard, Ensum, and Taylor developed the expected goals model (xG) based on logistic regression, which predicts the likelihood of a shot resulting in a goal\cite{pollard2004estimating}. Pollard and Reep\cite{pollard1997measuring} suggested estimating possessions in risk-reward terms by assigning a value of xG to shots taken during a possession and a value of -$xG$ to shots taken by the opponent team in the subsequent possession.

Spearman \cite{spearman2018beyond} introduced the term off-ball scoring opportunities (OBSO) which represents the posterior probability of scoring with the next on-ball event at a particular location based on tracking data. Spearman et al. proposed the idea of pitch control, based on a physical model that predicts that a pass will be accurate\cite{spearman2017physics}. Power et al. developed a pass model to analyze decision-making in risk-reward contexts\cite{power2017not}. 

Singh introduced the Expected Threat metric based on a Markov chain approach\cite{xt}. Cervone, D., D’Amour, A., Bornn, L., and Goldsberry suggested the expected possession value model (EPV) for basketball tracking data, taking into account individual player skills\cite{cervone2014pointwise}, \cite{cervone2016multiresolution}. Fernandez et al. apply EPV for soccer data\cite{fernandez2019decomposing}.

Garnier and Gregoir utilized deep reinforcement learning techniques and introduced a metric called Expected Discounted Goal (EDG)\cite{garnier2021evaluating}. Liu et al.\cite{liu2020deep} approached the estimation of EPV as a reinforcement learning (RL) task. They developed the Goal Impact Metric (GIM) as a result of their RL-based solution.

Dinsdale and Gallagher\cite{dinsdale2022transfer} predicted values of several player metrics for the first 1,000 minutes of a player at a new club, or the next 1,000 minutes if a player remains at their current club, taking into account Elo\cite{elo1967proposed} ratings of clubs and leagues.

\section{Proposed Approach for EPV}

\subsection{Background}

This section introduces our customized version of the Expected Possession Value (EPV) model. The model has been built using events data, which includes actions such as passes, shots, dribbles, etc. Each event is described by relevant details, including coordinates, time, and other information. This paper employs the following notation to formalize the problem:

\begin{itemize}
    \item $i$ - index of event,
    \item $c_i$ - control action,
    \item $d_i$ - duel action,
    \item $t_i$ - effective playing time of event $i$,
    \item $s_i$ - index of possession of event $i$,
    \item $xG_i$ - $xG$ of event $i$, if the event is a shot,
    \item $PV(e_i)$ - possession value of event $i$,
    \item $EPV(e_i)$ - expected possession value of event $i$ (prediction of EPV model).
\end{itemize}

By \textit{control actions} we consider the actions such as a pass, a shot, a dribble, a carry (ball control), a free kick, a goal kick, a penalty, a corner kick, a throw in. The idea behind this definition is that it is clear which player possesses the ball during these actions.

Aerial and ground duels are considered \textit{symmetrical duels}. Dribble is not a symmetrical duel.

We define \textit{possession} as a series of control actions by the same team uninterrupted by the opponent. In theory, team may touch the ball during the opponent's possession, like intercepting the ball, but if it doesn't lead to their control action, the opponent's possession continues. A goalkeeper's save, leading to a corner or offensive rebound, does not interrupt the opponent's possession. Therefore, one possession could include several shots.

\textit{Effective playing time} is the total amount of time that the ball is in play during the match. Most football metrics are normalized based on \textit{dirty} playing time. The duration of pauses during VAR reviews, substitutions, and set-piece situations should be excluded. Therefore, we will recalculate our metrics based on effective playing time, considering only the actual time of active gameplay. Furthermore, when we use terms like playing time, minutes, seconds, etc., we are referring to effective playing time.

\subsection{Possession Value of Control Actions}

We define our model's target variable as the \textit{possession value}. We then refer to the model's prediction as the \textit{expected possession value (EPV)}. Traditionally, there are two approaches to defining the EPV model's target. The first assigns a value of 1 to each event within a possession if it results in a goal, and 0 otherwise. However, we avoid this approach due to the limited number of goals, which could cause overfitting.

The alternative approach involves assigning each event within a possession the cumulative sum of $xG$ from \textit{future} shots during that possession.

\begin{equation}
PV(c_i) = (1 - \prod_{j=1}^{\infty} (1 - xG_j[t_j \geq t_i][s_i = s_j])) 
\end{equation}

We would like to highlight a couple of key points. Firstly, this formula is correct only to control actions, as exclusively in these cases can we definitively determine which player has possession. Secondly, we consider only the $xG$ of shots that occur after a specific event because the reward is determined by future events rather than past ones. Thirdly, we take into account the possibility of a subsequent goal occurring because the initial shot did not result in a goal. For that reason, the term $(1 - xG_j)$ is included in equation (1). A description of our \textit{xG} model can be found in the fourth chapter.

\subsection{Decay Effect}

Formula (1) has a significant drawback. It assigns the same possession value ($PV$) to all events preceding the corresponding shots, implying equal importance for each action. This assumption is invalid. Thus, we introduce a decay effect where actions leading to shots carry more weight. Hence, we modify the formula as follows:

\begin{equation}
PV(c_i) = (1 - \prod_{j=1}^{\infty} (1 - \gamma^{(t_j-t_i)}xG_j[t_j \geq t_i][s_i = s_j]) 
\end{equation}

The discount factor was proposed by Samuelson\cite{samuelson1938note} and has found widespread use in economics and reinforcement learning, prioritizing receiving rewards sooner. However, in our context, the interpretation of the discount factor differs. We don't encourage earlier shots. Instead, we suggest that if an action doesn't lead to a shot soon, it may not significantly advance the attacking phase. We use a discount factor of 0.95. This value is a hyperparameter that can be adjusted based on preferences. Analysts favoring a vertical attacking style might use a lower gamma value like 0.9, while those preferring a \textit{tiki-taka} style might choose a higher value like 0.99. 

Another point to highlight is that the formula operates correctly due to our consideration of effective playing time. For example, if a player draws a penalty, he receives a substantial reward since the penalty kick occurs immediately after. This ensures rewards accurately reflect the timing and impact of events in the game.

The concept of employing a discount factor in soccer is not entirely new. For example, in \cite{garnier2021evaluating}, the discounted expected number of goals that a team will score (or concede) was used as a target value for an RL algorithm. However, this target value lacks precision without taking into account the $xG$ and effective time concepts.

\subsection{Possession Risk}

Some $EPV$ models incorporate \textit{possession risk}\cite{pollard1997measuring}. An accurate forward pass not only increases the team's scoring chances but also reduces the threat near their own goal. These models consider the difference between the team's $PV$ and the opponent's $PV$ in the subsequent possession, rather than just the team's possession value.

The penalty for possession risk in this approach has a significant drawback. For instance, consider a player who completes a pass, but the team loses possession 10 seconds later, leading to the opponent earning a penalty kick 20 seconds after that. In this approach, the player's pass would be assigned a target value of -0.75 (the $xG$ value of the penalty kick), which is counterintuitive due to the elapsed time. To address this, we incorporate the decay effect from (2). Using the decay factor, the value is $-0.95^{30} * 0.75$ = -0.16, showing the player's action was less influential in the penalty.

Another issue with the original approach to assessing possession risk is that it only considers two possessions. However, a team might lose possession, quickly regain it, and score on the third possession. By incorporating the decay effect, we can account for a larger number of possessions.

Thus, we arrive at the following formula for the $PV$, which will be used as the target value for our \textit{EPV} model for control actions:

\begin{equation}
\begin{aligned}
PV(c_i) &= \sum_{s_j \in \text{team}} \left(1 - \prod_{j=1}^{\infty} (1 - \gamma^{(t_j - t_i)} xG_j[t_j \geq t_i])\right) \\
&\quad - \sum_{s_j \in \text{opponent}} \left(1 - \prod_{i=1}^{\infty} (1 - \gamma^{(t_j - t_i)} xG_j[t_j \geq t_i])\right)
\end{aligned}
\end{equation}

\subsection{Possession Value of Symmetrical Duels}

We also calculate \textit{EPV} for symmetrical duels. To address the challenge of assigning possession for duel events, we assign the possession value of the first control action following the duel to that duel. If there is a series of symmetrical duels, all are recursively assigned the same possession value.

\begin{equation}
  PV(d_i) =
    \begin{cases}
      PV(e_{i+1}) & \text{if } s_i = s_{i+1} \\
      - PV(e_{i+1}) & \text{if } s_i \neq s_{i+1} \\
    \end{cases}       
\end{equation}

\subsection{Reward Metrics}

Before building the \textit{EPV} model, we define our research goal. Our objective is to implement a metric that measures a player/team's ability to "improve" possession. We highlight five possible outcomes of a player's actions:

\begin{enumerate}
    \item The action leads to a control action by the same team.
    \item The action leads to a control action by the opponent team.
    \item The goal was scored.
    \item The action was the last in the half.
    \item The action leads to a symmetrical duel.
\end{enumerate}

In the first scenario, if a control action or symmetrical duel maintains possession, the player receives a \textit{reward} equal to the difference between the $EPV$ values of neighboring events. In the second scenario, if a control action or duel results in a turnover, the next ($i+1$) event is by the opposing team. Thus, the player is penalized twice: once for the turnover and once for providing a goal opportunity to the opponent.

If possession ends with a goal, the player receives a reward of ($1 - EPV(c_i) - EPV(c_{i+1})$), where $c_{i+1}$ denotes a pass from the center circle after a goal. In the fourth scenario, we assign a reward value of zero to neither reward nor punish the player.

The fifth scenario is the most challenging. For example, a pass into an aerial duel initially gets a negative reward for creating a 50/50 situation. However, we must consider the potential mismatch between players involved. We address this by using an improved approach to estimate symmetrical duel skills \cite{shelopugin2023evaluating}. By adding the probability of winning a duel as a feature in the EPV model, we can more accurately evaluate the reward. If the teammate has a high probability of winning the duel, the passing player receives a greater reward. However, when calculating the reward for players in symmetrical duels, we won't consider the probability of winning the duel. Instead, we compare the actual outcome with the "average" outcome to avoid penalizing a player for superior skills by inflating the EPV.

Different feature sets describe control actions and symmetrical duels. Hence, we built three types of models: one for control actions, one for symmetrical duels for the "average" player, and one for symmetrical duels with player duel skill estimation. These are denoted as $EPV$, $EPV_{duel}^{avg}$, and $EPV_{duel}^{ind}$, respectively. Thus, we've designed this $reward$ formula to evaluate players' control actions and symmetrical duels.

If $i$ is a control action:

\begin{equation}
\Delta EPV(c_i) =
\begin{cases}

    EPV(c_{i+1}) - EPV(c_i), & \text{if scenario 1} \\
    
    - EPV(c_{i+1}) - EPV(c_i), & \text{if scenario 2}  \\

    1 - EPV(c_i) - EPV(c_{i+1}), & \text{if scenario 3}  \\
    
    0, & \text{if scenario 4}  \\
    
    EPV_{duel}^{ind}(d_{i+1}) - EPV(c_i), & \text{if scenario 5} \\

\end{cases}
\end{equation}

If $i$ is a symmetrical duel:

\begin{equation}
\Delta EPV(d_i) =
\begin{cases}

    EPV(c_{i+1}) - EPV_{duel}^{avg}(d_i), & \text{if scenario 1} \\
    
    - EPV(c_{i+1}) - EPV_{duel}^{avg}(d_i), & \text{if scenario 2}  \\
    
    0, & \text{if scenario 4}  \\
    
    EPV_{duel}^{ind}(d_{i+1}) - EPV_{duel}^{avg}(d_i), & \text{if scenario 5} \\

\end{cases}
\end{equation}

Regarding our target function, it's crucial to understand that pass accuracy doesn't directly affect rewards. For instance, an inaccurate cross leading to a defender's handball still results in a penalty kick attempt. This underscores the use of the concept of \textit{possession} in our approach. Additionally, our analysis excludes certain events like interceptions. Hence, when calculating rewards, we omit these actions. For instance, a sequence like "pass-interception-shot" is interpreted as "pass-shot" in our analysis.

\section{EPV Implementation}

\subsection{Expected Goals}
Soccer's low-scoring nature, averaging 2.6 goals per game, poses challenges in assessing player performance. The emergence of $xG$ as a measure of scoring opportunities addresses this challenge, representing the probability of a shot resulting in a goal, solely based on game situation, not individual player skills. Our goal is to develop a model that evaluates scoring opportunities for the average player.

We create two distinct models: one for set-piece shots and another for open-play shots. Factors such as shot coordinates, distance, angle from the goal, and set-piece type (penalty, free-kick, corner) were considered. For open-play shots, attributes like body part used and preceding actions (e.g., aerial duel, pass) were included, along with spatial details. Separate model was built for set-pieces, as $xG$ should be independent of previous events.

Certain features like current game score, championship, and player's team were omitted due to their correlation with player skill. Another challenge is the dataset revealed a non-uniform distribution of shots, with top players having more scoring opportunities. To tackle this, we implemented a custom loss function, choosing a log-loss function as the basic loss function for each shot. We divided the loss function value for each shot by its appearance count in the training set, reducing the focus on overrepresented players.

\begin{equation}
customlogloss_{i} = \frac{1}{|player_{i} \in D|} [y_i \log(p_i) + (1 - y_i) \log(1 - p_i)]
\end{equation}

We trained the LightGBM \cite{ke2017lightgbm} and CatBoost \cite{prokhorenkova2018catboost} frameworks, with LightGBM showing superior accuracy. We implemented equation (7) as an \textit{objective} parameter in this framework.

\subsection{Symmetrical Duels}

Here, we present an improved version of our approach \cite{shelopugin2023evaluating}. However, we focus only on \textit{symmetrical duels}.

Analysts often use win percentage as a measure of player duel skill, but this overlooks opponent strength. Players may have high win rates against weaker opponents. Additionally, coaches often match players of similar strength, especially in set-pieces, leading to win rates converging to fifty percent as strong players face strong opponents and weak players face weak ones.

Another approach suggested by Garry Gelade \cite{gelade} uses the Bradley-Terry model \cite{bradley1952rank} to calculate player ratings for duels. This approach is superior as it considers opponent strength and enables modeling of future situations. The metric is transferable across leagues, allowing comparisons between players in diverse competitions.

Gelade's method has limitations. It assumes that players have equal chances to win duels without considering external factors, which isn't always accurate. For example, in aerial duels, defending players have an advantage as they face the opponent's goal during the defensive phase. Another limitation is that the Bradley-Terry model is not state-of-the-art. Therefore, we opted to use a modified version of Glicko-2 \cite{glickman2012example}.

Introduce a definition of a duel winner. In both cases (aerial and ground duels), we will adhere to the following logic:

\begin{enumerate}
    \item If a player suffered a foul, he is considered the winner.
    \item A player who makes the first touch on the ball is considered the winner.
    \item If no one touches the ball, we will consider the player whose team gains possession after the duel as the winner.
\end{enumerate}

We must consider that duels aren't fully symmetrical. Outcomes depend not only on players' skills but also on external factors, such as the location of the duel or the type of pass leading to it. 

The original Glicko-2 version updates rating in this way:

\begin{equation}
    \mu^{'} = \mu + \phi^{'2}g(\phi_j)(s_j - E(\mu, \mu_j, \phi_j))
\end{equation}

We have modified it for a defender: 

\begin{equation}
\mu^{'} = \mu + \phi^{'2}g(\phi_j)(s_j - E(\mu + a, \mu_j, \phi_j)) 
\end{equation}

We determine advantage as follows: using a model predicting duel outcomes based on contextual features like duel and pass coordinates, pass type (e.g., corner, free kick), and number of opponents. We exclude player skill-related features, aiming to create an "average" model describing duel difficulty.

We encounter a data leak due to two factors. Firstly, the defending team holds an advantage in aerial duels. Secondly, central defenders typically excel in aerial duels compared to other positions, but they occur more frequently in defensive roles, leading to underestimated ratings. To address this, we categorize player positions into six groups: central defenders, full-backs, midfielders, central forwards, wingers, and goalkeepers. We train model by filtering aerial duels where players involved occupy the same position category. We apply the same logic as described in formula (7).

Therefore, we train a LightGBM model to calculate the probability of winning a duel. Based on this probability, we calculate the average advantage ($a$ from (9)) for a Glicko-2-based model. This allows us to determine individual aerial and ground duel ratings for each player, considering the duel context. The ratings are presented in Tables \ref{aerials_ratings} and \ref{grounds_ratings}.

\subsection{Expected Possession Value}

We built the EPV model in a similar way to the $xG$ model, utilizing spatial characteristics of the action and the preceding one. Six separate models were trained:

    1. For control actions in open-player situations.
    
    2. For set pieces, as they should depend only on the current action.
    
    3-4. Aerial and ground average EPV models, $EPV_{duel}^{avg}$. These models describe the context of the duel while ignoring the player's skill in aerial duels. We used the predictions of the LightGBM model from chapter 4.2 as a feature to describe the context of the duel. These models help calculate the reward (6), which allows to estimate players' abilities to "improve" possession in situations where the player participates in a duel.
    
    5-6. In contrast to the previous point, here we want to describe the aerial or ground duel situation considering player skills. It helps calculate pass reward more accurately, as seen in scenario 5 of formulas (5), (6). We used the following features: the probability to win the duel (taking into account player ratings), the player's rating, the opponent's rating, as well as the spatial features of the duel and the pass leading to the duel.

Analogously to equation (7), we implemented a custom mean squared error:

\begin{equation}
customMSE_{i} = \frac{1}{|player_{i} \in D|} (y_i - \hat{y}_i)^2
\end{equation}

\section{Metrics Prediction for the Upcoming Season}

\subsection{Season Pass Carry Reward}

We introduce the metric \textit{Pass Carry Reward} ($PCR$) as a measure of how a player enhances possession through passes or carries over the season. For the term \textit{carry}, we encompass all events where a player controls the ball, including dribbles, carries, or simply maintaining possession without significant movement. By summing all $\Delta EPV(e_i)$ values throughout the season, we can derive a season reward, designated as $PCR$, focusing solely on pass and carry events. To assess a player's season performance accurately, it's crucial to normalize this reward based on the player's effective playing time, specifically their effective time per 60 minutes of play.

\begin{align}
PCR(player) = \frac{60 \times \sum \Delta EPV(e_i|player, e_i=pass \lor  e_i=carry)}{\sum \text{minutes}}
\end{align}

The idea behind this paper is straightforward: to predict players' $PCR$ for the upcoming season. However, there are certain limitations and restrictions that we need to address and discuss.

\subsection{Training Set}

First and foremost, it is important to define the group of players for whom we will train our model. Our decision is to focus only on players who have accumulated at least 100 minutes in both the current season and the next season. This criterion is implemented to ensure the stability of $PCR$ predictions, as values of $PCR$ for players with limited playing time tend to be less reliable. The potential drawbacks of dropping data based on this criterion will be discussed in the subsequent subsection.

\subsection{Features}

The calculated features, approximately 600 in total, can be categorized into several groups:

\textbf{Player-specific features.} These include attributes such as age, height, position, and other characteristics that describe the individual player.

\textbf{Performance features.} This group encompasses metrics like $PCR$, $xG$, goals, played minutes, and other raw statistics from the previous seasons. Additionally, average values of these metrics over the past three or five seasons are also considered.

\textbf{Contribution to team success.} These statistics capture the player's impact on the team's overall performance. For instance, features like the player's share of the team's total $xG$ when they were on the field are included. These features are crucial as they provide insights into a player's effectiveness, especially when playing for a weaker team where their absolute metrics may be lower.

\textbf{League style features.} This category includes metrics that describe the prevalent playing style in the league. For example, the average $PCR$ of the league in the previous season is considered. These features help address challenges such as the influence of league-specific playing styles on $PCR$. For a player in a league that prioritizes attacking play, it may be easier to achieve higher $PCR$ values.

\textbf{Team and league strength features.} These features account for factors such as team and league strength, particularly in the context of player transfers. They provide information regarding a player's transfer, such as moving from a strong team to a weaker one, and help contextualize the player's performance accordingly. A detailed discussion regarding this last group of features will be presented in the subsequent subsection.

\subsection{Ratings of Clubs and Leagues}

To address the issue of player transfers and the potential impact on $PCR$ prediction, it is important to consider the information associated with changes in clubs or leagues. One approach could involve introducing a categorical feature that indicates the transfer from one league to another, such as "player transfers from Denmark Superliga to France Ligue 1". However, this approach may encounter challenges related to the curse of dimensionality, as there may not be a significant number of transfers between specific leagues. Additionally, it is worth noting that the strengths of leagues can vary over time.

To overcome these challenges, we propose utilizing a modified Glicko-2 rating system \cite{shelopugin2023ratings}. This model calculates ratings for teams based on match outcomes, allowing for the estimation of team strength. The strength of a league is defined as the average rating of a specified number of teams, with the exact number being adjustable to account for variations across leagues.

With this modified rating system, we can calculate various features, such as the rating of the player's old team, the rating of the player's new team, and the difference in ratings between the player's old league and the new league. These features provide valuable information related to the player's transition and enable us to account for changes in team and league strength. Additionally, we calculate the average rating of team opponents, as two players can demonstrate similar metrics; however, the level of opponents can differ.

It is important to note that we utilized the actual club/league ratings at the start of the season. This approach allows the model to incorporate up-to-date information about the strength of the clubs/leagues, considering that their levels may change over time. Moreover, we must adhere to the natural constraint that we cannot use the future ratings of the clubs/leagues.

\subsection{Probability to Stay in the Data}

As mentioned earlier, our training set includes only instances where players have played at least 100 minutes in the next season. This limitation poses a challenge as it is not possible to know in advance how many minutes a player will play in the next season. To address this issue, we build an auxiliary model.

We predict whether a football player will play at least 100 minutes in the next season. A player may have less than 100 minutes for various reasons: retirement from professional football, injury, underperformance resulting in exclusion from matches by the coach, or transfer to a team or league not presented in the dataset, indicating the transfer to a less competitive league.

We train the model using the same features as those employed in the previous model, excluding the features related to the ratings of a new team or league. In addition, we incorporated data on players' contract durations at the beginning of the season, leveraging information sourced from the FIFA video game series\cite{fifa}.

\subsection{Presence-only Data Problem}

Another challenge in training our model is the non-random nature of real-world transfers. Football club managers make informed decisions when acquiring players, leading to two key scenarios. When a player moves from a weaker club to a stronger one, it suggests the new club sees potential or talent in the player. When a player moves from a strong club to a weaker one, it suggests a decline in performance. These scenarios introduce bias in our predictions. We use features like the "rating of the new team" to inform the model about transfers, but the outcomes remain uncertain. Consequently, predictions can be overly optimistic in the first scenario and overly pessimistic in the second, reflecting the uncertainties and biases inherent in real-world transfers.

This problem is known as \textit{presence-only data} or \textit{learning from positive and unlabeled data} \cite{ward2009presence}. First, account for the data leak from leaving data, assigning a probability $pl$. Second, consider the data leak from subsequent transfers, which is highly correlated with league rating differences. Assign this as $\Delta ratings$, with 1500 as the average club rating.

\begin{align}
\Delta ratings = \frac{(rating(league_{new}) - rating(league_{old}))}{1500}
\end{align}

\begin{align}
PCR_{adj} = PCR * 0.8 ^ {(\Delta ratings + pl)}
\end{align}

\section{Results}

As a baseline, we calculate the root mean squared error ($RMSE$) and mean absolute error ($MAE$) between the $PCR$ from the previous season and the next season (refer to Table \ref{metrics_decomposition_without_ml}). This imitates a selection process based on historical data without applying machine learning. We separate the data into several groups based on the following binary rules: whether the player is in the same team/league as in the previous season or not, and whether the player played more than 1000 minutes in the previous season or not. We remember that we are trying to solve a selection problem; therefore, we assume that a player may change clubs or even leagues.

\begin{table}[htbp]
    \centering
    \small
    \caption{Baseline Metrics Without Applying Machine Learning}
        \begin{tabular}{|l|l|l|l|l|}
    \hline
        Data Sample & RMSE, >100 min & MAE, >100 min & RMSE, >1000 min & MAE, >1000 min \\ \hline
        all data & 0.053 & 0.036 & 0.042 & 0.029 \\ \hline
        the same team, the same league & 0.05 & 0.034 & 0.039 & 0.027 \\ \hline
        the same team, a new league & 0.051 & 0.035 & 0.042 & 0.031 \\ \hline
        a new team, the same league & 0.055 & 0.038 & 0.044 & 0.031 \\ \hline
        a new team, a new league & 0.061 & 0.043 & 0.051 & 0.036 \\ \hline
    \end{tabular}
    \label{metrics_decomposition_without_ml}
\end{table}

By comparing the RMSE and MAE with the predictions of our model (without adjustment), we can estimate the contribution of our approach (refer to Table \ref{metrics_decomposition}). 

\begin{table}[htbp]
    \centering
    \small
    \caption{Achieved Metrics Based On Our Approach}
        \begin{tabular}{|l|l|l|l|l|}
    \hline
        Data Sample & RMSE, >100 min & MAE, >100 min & RMSE, >1000 min & MAE, >1000 min \\ \hline
        all data & 0.033 & 0.023 & 0.031 & 0.021 \\ \hline
        the same team, the same league & 0.032 & 0.022 & 0.029 & 0.02 \\ \hline
        the same team, a new league & 0.031 & 0.021 & 0.027 & 0.019 \\ \hline
        a new team, the same league & 0.034 & 0.024 & 0.032 & 0.023 \\ \hline
        a new team, a new league & 0.037 & 0.026 & 0.036 & 0.025 \\ \hline
    \end{tabular}
    \label{metrics_decomposition}
\end{table}

However, the above-mentioned evaluation method has its limitations. Our end product is a player ranking based on predicted $PCR$, and there is no guarantee that the calculated shortlist is accurate. Additionally, there is no mathematical way to prove that $EPV$-based metrics truly correlate with player skills. We see two ways to evaluate our shortlists. The first is to ask experts, which may be done in future work. The second way is to assume that the transfer market is enough effective at least at the top level and compare top players from our shortlists with actual transfers to top clubs.

Table \ref{historical_players_epv} presents the top season performances based on the $PCR$ metric among players who played at least 1,000 minutes during the season. The $PCR$ predictions for Manchester City, Barcelona, Milan, and Brighton can be found in Tables 7-10. Brighton was chosen because this club has a reputation for having a strong data analysis department and making transfers based on data. The actual date is June 1, 2024.

There is also a case study section in the appendix that demonstrates the advantage of using individual duel skills in the $EPV$ model.


\section{Conclusion}\label{sec2}

We improve the EPV model by accurately accounting for effective playing time, decay effect, and possession risk. The incorporation of player duel skills allows us to calculate rewards more accurately. Therefore, we partially solve the problem of reward distribution between the passing player and the target player in cases of passes leading to duels. However, the problem still exists in cases of accurate passes, and we do not have a solution based on event data.

This model can potentially demonstrate higher performance with tracking data due to the consideration of more features. The inclusion of injury data is promising not only for enhancing the predictive capacity of the model that calculates the probability of a player remaining in the dataset but also for improving the prediction of $PCR$ for the next season. Additionally, we need to consider the context of new and old teams more accurately. For instance, we can replace $PCR$ with a conditional expectation that takes into account the player's position in the next club. For example, our model lacks accuracy when a center forward starts to play as a winger in a new club. Finally, the presence-only data problem should be tackled with a more elegant approach, which is a subject for further research.

\section{Acknowledgments}

The author is thankful to Nikita Kozodoi, Viktoria Lokteva, Nikita Vasyukhin, Iskander Safiulin, Daniil Babaev, and Alexander Sirotkin for their invaluable consultations on machine learning and soccer analytics.

\bibliographystyle{IEEEtran}
\bibliography{EPV}

\newpage

\appendix


\section*{Appendix}

\section{A Case Study: Gianluigi Donnarumma Passes to Duels in Seasons 19/20, 20/21} 

In this paper, we focus on solving the selection problem. The main idea of the work is to suggest a mechanism to narrow down the selection pool. However, our approach can find applications not only in selection tasks but also in team performance and opposition analysis. Below, we provide an example of such an analysis.

Gianluigi Donnarumma was a great goalkeeper in terms of shot-stopping during his career at AC Milan. However, he had some struggles with his passes. Coach Stefano Pioli suggested the following solution: if Donnarumma was under pressure, he would make a long forward pass (usually leading to a duel); otherwise, he would make a short pass to a defender or defensive midfielder.

By long forward pass, we define a pass with a distance of more than 40 metres and an increment of y-projection of more than 10 metres. The most popular target players for Donnarumma were Zlatan Ibrahimovic (35 aerial duels, 8 ground duels) and Rafael Leao (28 aerial duels, 11 ground duels).

The traditional EPV approach does not take into account players' duel skills; however, our approach provides a more detailed analysis. In Table \ref{CaseStudyDonnarumma}, we compare passes to Leao and Ibrahimovic. The column $saved$ represents the proportion of possessions that were saved after duels. The column $apriori$ represents the probability of winning a duel without considering a player's skills, and the column $win\_duel$ represents the probability of winning a duel while accounting for duel skills. As we can see, Leao and Ibrahimovic were in similar situations in terms of the difficulty of winning duels. However, despite facing slightly stronger opponents (as indicated by $opp\_rating$), Zlatan managed to convert these situations into winning scenarios: his average probability of winning was 61.1\%, compared to Leao's 35.8\% in aerial duels. As a result, Ibrahimovic's $EPV$ (column $epv\_ind\_duel$) values are higher than Leao's. Take note that the Glicko ratings have a property to update over time; therefore, the column $rating$ represents the average ratings of Ibrahimovic and Leao against their opponents after Donnarumma's passes.

\begin{table}[htbp]
    \centering
    \small
    \caption{Gianluigi Donnarumma Passes to Rafael Leao or Zlatan Ibrahimovic Duels}
    \begin{tabular}{|l|l|l|l|l|l|l|l|l|l|}
\hline
    player & duel & duels & saved & apriori & win\_duel & rating & opp\_rating & duel\_epv & epv\_ind\_duel \\ \hline
    Leão & aerial & 35 & 28.6 & 40.5 & 35.8 & 1519 & 1545 & 0.00075 & 0.00069 \\ \hline
    Leão & ground & 8 & 12.5 & 45.0 & 28.2 & 1408 & 1534 & 0.00073 & 0.00062 \\ \hline
    Zlatan & aerial & 28 & 53.6 & 39.2 & 61.1 & 1746 & 1573 & 0.00077 & 0.00135 \\ \hline
    Zlatan & ground & 11 & 54.5 & 44.8 & 44.9 & 1539 & 1533 & -0.00012 & -0.0001 \\ \hline
\end{tabular}
    \label{CaseStudyDonnarumma}
\end{table}

\clearpage

\begin{table}[H]
    \centering
    \caption{Aerial Duel Ratings}
    \begin{tabular}{|l|l|l|l|l|l|}
    \hline
        \# & player & position & rating & \#duels & wins\% \\ \hline
        1 & V. van Dijk & central\_def & 1762 & 2167 & 71.9 \\ \hline
        2 & L. Sobiech & central\_def & 1761 & 1239 & 72.2 \\ \hline
        3 & B. Matić & midfielder & 1757 & 1076 & 71.1 \\ \hline
        4 & M. Djurić & central\_forward & 1744 & 2763 & 67.7 \\ \hline
        5 & L. Došek & central\_forward & 1743 & 253 & 66.0 \\ \hline
        6 & S. Coates & central\_def & 1742 & 1729 & 69.3 \\ \hline
        7 & Rafael Donato & central\_def & 1742 & 868 & 71.5 \\ \hline
        8 & P. Budkivskyi & central\_forward & 1738 & 1238 & 56.2 \\ \hline
        9 & A. Papazoglou & wing & 1734 & 1137 & 61.3 \\ \hline
        10 & M. Jakubko & central\_forward & 1726 & 239 & 63.6 \\ \hline
        11 & S. Memišević & central\_def & 1725 & 762 & 67.2 \\ \hline
        12 & N. Bendtner & central\_forward & 1725 & 742 & 52.4 \\ \hline
        13 & S. Ganvoula & central\_forward & 1722 & 1364 & 58.7 \\ \hline
        14 & J. Tarkowski & central\_def & 1721 & 2097 & 66.3 \\ \hline
        15 & R. Uldriķis & central\_forward & 1721 & 1615 & 56.6 \\ \hline
        16 & C. Bolger & central\_def & 1715 & 670 & 74.0 \\ \hline
        17 & F. Benković & central\_def & 1715 & 690 & 77.2 \\ \hline
        18 & A. Flint & central\_def & 1713 & 2676 & 68.8 \\ \hline
        19 & R. Rodelin & wing & 1713 & 1461 & 65.6 \\ \hline
        20 & K. Moore & central\_forward & 1713 & 2478 & 58.2 \\ \hline
        21 & C. Keeley & central\_def & 1712 & 687 & 72.6 \\ \hline
        22 & J. Greaves & lateral & 1712 & 774 & 71.2 \\ \hline
        23 & J. Cooper & central\_def & 1712 & 2915 & 69.6 \\ \hline
        24 & R. Leeuwin & central\_def & 1711 & 754 & 67.8 \\ \hline
        25 & L. Lindsay & central\_def & 1711 & 1732 & 68.7 \\ \hline
        26 & E. Sipović & central\_def & 1711 & 266 & 76.3 \\ \hline
        27 & M. Papadopulos & central\_forward & 1710 & 1807 & 55.2 \\ \hline
        28 & J. Owens & central\_forward & 1709 & 1590 & 59.7 \\ \hline
        29 & C. Casadei & wing & 1708 & 106 & 70.8 \\ \hline
        30 & V. Posmac & central\_def & 1707 & 870 & 77.2 \\ \hline
        31 & J. Arango & wing & 1706 & 165 & 64.2 \\ \hline
        32 & F. Mayembo & central\_def & 1705 & 446 & 69.7 \\ \hline
        33 & R. Sykes & central\_def & 1705 & 195 & 73.3 \\ \hline
        34 & V. Koskimaa & central\_def & 1704 & 596 & 74.8 \\ \hline
        35 & Y. Khacheridi & central\_def & 1704 & 532 & 76.3 \\ \hline
        36 & N. Ziabaris & central\_def & 1704 & 262 & 72.9 \\ \hline
        37 & I. Soumaré & wing & 1703 & 684 & 61.3 \\ \hline
        38 & C. Stewart & central\_def & 1702 & 458 & 63.3 \\ \hline
        39 & J. Chabot & central\_def & 1702 & 808 & 63.2 \\ \hline
        40 & C. Bocșan & central\_def & 1700 & 235 & 71.1 \\ \hline
        41 & J. Blayac & central\_forward & 1700 & 720 & 59.9 \\ \hline
        42 & Héliton & central\_def & 1700 & 360 & 70.8 \\ \hline
        43 & Nilton & midfielder & 1699 & 356 & 64.0 \\ \hline
        44 & M. Janko & central\_forward & 1699 & 647 & 59.7 \\ \hline
        45 & M. Baša & central\_def & 1699 & 242 & 76.9 \\ \hline
        46 & D. Lemajić & central\_forward & 1699 & 1420 & 54.7 \\ \hline
        47 & Kim Min-Jae & central\_def & 1699 & 485 & 67.2 \\ \hline
        48 & A. Taylor & central\_def & 1699 & 1353 & 71.2 \\ \hline
        49 & M. Beevers & central\_def & 1698 & 787 & 67.0 \\ \hline
        50 & A. Zabolotny & central\_forward & 1698 & 1572 & 57.8 \\ \hline
    \end{tabular}
    \label{aerials_ratings}
\end{table}

\begin{table}[H]
    \centering
    \caption{Ground Duel Ratings}
        \begin{tabular}{|l|l|l|l|l|l|}
    \hline
        \# & player & position & rating & \#duels & wins\% \\ \hline
        1 & B. Ostojić & central\_def & 1695 & 279 & 73.1 \\ \hline
        2 & Felipe & central\_def & 1672 & 643 & 67.8 \\ \hline
        3 & Vasco Fernandes & central\_def & 1671 & 408 & 66.4 \\ \hline
        4 & V. van Dijk & central\_def & 1669 & 742 & 71.6 \\ \hline
        5 & R. García & central\_def & 1666 & 427 & 68.1 \\ \hline
        6 & L. Valenti & central\_def & 1666 & 135 & 74.1 \\ \hline
        7 & V. Mudrac & central\_def & 1665 & 274 & 69.3 \\ \hline
        8 & M. Vítor & central\_def & 1663 & 691 & 68.7 \\ \hline
        9 & Raúl Navas & central\_def & 1663 & 350 & 70.9 \\ \hline
        10 & N. Elvedi & central\_def & 1660 & 735 & 66.9 \\ \hline
        11 & L. McCullough & central\_def & 1658 & 317 & 75.7 \\ \hline
        12 & G. Valsvik & central\_def & 1656 & 702 & 64.2 \\ \hline
        13 & P. Jagielka & central\_def & 1656 & 388 & 69.8 \\ \hline
        14 & J. Matip & central\_def & 1655 & 543 & 69.2 \\ \hline
        15 & A. Maxsø & central\_def & 1655 & 552 & 65.4 \\ \hline
        16 & A. Flint & central\_def & 1655 & 917 & 68.5 \\ \hline
        17 & J. Kucka & midfielder & 1654 & 877 & 55.6 \\ \hline
        18 & Marcão & central\_def & 1654 & 427 & 65.6 \\ \hline
        19 & A. Bengtsson & lateral & 1654 & 118 & 72.0 \\ \hline
        20 & Rodrigão & central\_def & 1654 & 293 & 68.9 \\ \hline
        21 & Aleksandar Luković & central\_def & 1653 & 120 & 76.7 \\ \hline
        22 & T. Petrášek & central\_def & 1653 & 254 & 72.0 \\ \hline
        23 & C. McJannet & central\_def & 1653 & 445 & 67.2 \\ \hline
        24 & N. Belaković & wing & 1652 & 351 & 61.5 \\ \hline
        25 & Gabriel Magalhaes & central\_def & 1651 & 566 & 67.3 \\ \hline
        26 & M. Hasebe & midfielder & 1651 & 572 & 63.5 \\ \hline
        27 & M. Peersman & lateral & 1651 & 817 & 62.5 \\ \hline
        28 & G. Mancini & central\_def & 1650 & 759 & 63.9 \\ \hline
        29 & O. Bačo & central\_def & 1648 & 478 & 61.3 \\ \hline
        30 & H. Moukoudi & central\_def & 1647 & 408 & 61.3 \\ \hline
        31 & L. Querfeld & central\_def & 1647 & 330 & 67.9 \\ \hline
        32 & C. Halkett & central\_def & 1646 & 374 & 68.2 \\ \hline
        33 & W. Pacho & central\_def & 1646 & 269 & 66.2 \\ \hline
        34 & S. Bell & central\_def & 1646 & 543 & 61.7 \\ \hline
        35 & Luiz Otávio & central\_def & 1645 & 344 & 68.0 \\ \hline
        36 & B. Utvik & central\_def & 1644 & 398 & 64.3 \\ \hline
        37 & S. Papagiannopoulos & central\_def & 1644 & 716 & 70.4 \\ \hline
        38 & L. Nielsen & central\_def & 1644 & 708 & 65.5 \\ \hline
        39 & R. Grodzicki & central\_def & 1644 & 171 & 70.2 \\ \hline
        40 & A. Sørensen & central\_def & 1644 & 589 & 65.9 \\ \hline
        41 & C. Goldson & central\_def & 1644 & 820 & 72.8 \\ \hline
        42 & F. Aguilar & central\_def & 1644 & 117 & 72.6 \\ \hline
        43 & O. Kúdela & central\_def & 1644 & 465 & 66.2 \\ \hline
        44 & J. Boller & central\_def & 1643 & 280 & 68.2 \\ \hline
        45 & M. Lovato & central\_def & 1643 & 267 & 69.7 \\ \hline
        46 & L. Morgillo & central\_def & 1643 & 130 & 73.1 \\ \hline
        47 & Tiago Ilori & central\_def & 1643 & 200 & 67.5 \\ \hline
        48 & M. Connolly & central\_def & 1642 & 380 & 69.7 \\ \hline
        49 & T. Vion & lateral & 1642 & 419 & 59.4 \\ \hline
        50 & A. Seck & central\_def & 1642 & 444 & 68.7 \\ \hline
    \end{tabular}
    \label{grounds_ratings}
\end{table}

\begin{table}[H]
    \centering
    \caption{Top Season Performances Based On PCR}
    \begin{tabular}{|l|l|l|l|l|l|l|}
    \hline
        \# & player & team & competition & season & PCR & eff\_time \\ \hline
        1 & H. Ziyech & Ajax & Netherlands. First & 2019/2020 & 0.355 & 1237 \\ \hline
        2 & J. Iličić & Atalanta & Italy. First & 2020/2021 & 0.315 & 1143 \\ \hline
        3 & A. Zoubir & Qarabag & Azerbaijan. First & 2020/2021 & 0.314 & 1465 \\ \hline
        4 & H. Ziyech & Ajax & Netherlands. First & 2017/2018 & 0.308 & 2257 \\ \hline
        5 & O. Dembélé & Barcelona & Spain. First & 2021/2022 & 0.305 & 1008 \\ \hline
        6 & J. Hauge & Bodo Glimt & Norway. First & 2020 & 0.305 & 1149 \\ \hline
        7 & J. Grealish & Aston Villa & England. First & 2020/2021 & 0.301 & 1481 \\ \hline
        8 & J. Doku & Manchester City & England. First & 2023/2024 & 0.299 & 1207 \\ \hline
        9 & Malcom & Zenit & Russia. First & 2020/2021 & 0.297 & 1184 \\ \hline
        10 & A. Zoubir & Qarabag & Azerbaijan. First & 2023/2024 & 0.287 & 1134 \\ \hline
        11 & A. Schjelderup & Nordsjaelland & Denmark. First & 2023/2024 & 0.283 & 1207 \\ \hline
        12 & L. Messi & Barcelona & Spain. First & 2018/2019 & 0.277 & 2058 \\ \hline
        13 & T. Ali & Malmo FF & Sweden. First & 2023 & 0.277 & 1258 \\ \hline
        14 & K. Mbappé & PSG & France. First & 2019/2020 & 0.276 & 1146 \\ \hline
        15 & J. Ito & Genk & Belgium. First & 2021/2022 & 0.273 & 2286 \\ \hline
        16 & Neymar & PSG & France. First & 2017/2018 & 0.272 & 1353 \\ \hline
        17 & L. Abada & Celtic & Scotland. First & 2022/2023 & 0.271 & 1078 \\ \hline
        18 & Y. Soteldo & Santos & Brazil. First & 2023 & 0.27 & 1105 \\ \hline
        19 & Musa Al Tamari & APOEL & Cyprus. First & 2018/2019 & 0.27 & 1387 \\ \hline
        20 & J. Clarke & Sunderland & England. Second & 2023/2024 & 0.268 & 2503 \\ \hline
        21 & R. Mahrez & Manchester City & England. First & 2019/2020 & 0.268 & 1506 \\ \hline
        22 & G. Kanga & Crvena Zvezda & Serbia. First & 2016/2017 & 0.268 & 1261 \\ \hline
        23 & M. Marin & Crvena Zvezda & Serbia. First & 2018/2019 & 0.262 & 1145 \\ \hline
        24 & Dodo & Hamrun Spartans & Malta. First & 2020/2021 & 0.258 & 1033 \\ \hline
        25 & P. Zinckernagel & Bodo Glimt & Norway. First & 2020 & 0.254 & 1818 \\ \hline
        26 & Daniel Podence & Olympiacos Piraeus & Greece. First & 2023/2024 & 0.252 & 1199 \\ \hline
        27 & C. McCloskey & Glenavon & Northern Ireland. First & 2020/2021 & 0.251 & 1107 \\ \hline
        28 & K. Mbappé & Monaco & France. First & 2016/2017 & 0.25 & 1088 \\ \hline
        29 & J. Cooper & Linfield & Northern Ireland. First & 2019/2020 & 0.249 & 1463 \\ \hline
        30 & A. Limbombe & Almere City & Netherlands. Second & 2022/2023 & 0.248 & 1360 \\ \hline
        31 & I. Kallon & Cambuur & Netherlands. Second & 2020/2021 & 0.248 & 1530 \\ \hline
        32 & Antony & Ajax & Netherlands. First & 2021/2022 & 0.244 & 1323 \\ \hline
        33 & N. Lang & Club Brugge & Belgium. First & 2022/2023 & 0.242 & 1646 \\ \hline
        34 & A. Abreu & UT Petange & Luxembourg. First & 2022/2023 & 0.241 & 1437 \\ \hline
        35 & D. Payet & Olympique Marseille & France. First & 2017/2018 & 0.241 & 1660 \\ \hline
        36 & Ângelo Gabriel & Santos & Brazil. First & 2022 & 0.24 & 1085 \\ \hline
        37 & H. Ziyech & Ajax & Netherlands. First & 2018/2019 & 0.24 & 1794 \\ \hline
        38 & J. Croux & Roda JC & Netherlands. Second & 2019/2020 & 0.236 & 1856 \\ \hline
        39 & L. Messi & Barcelona & Spain. First & 2015/2016 & 0.236 & 2025 \\ \hline
        40 & Á. Di María & PSG & France. First & 2020/2021 & 0.235 & 1372 \\ \hline
        41 & Marquinhos & Ferencvaros & Hungary. First & 2023/2024 & 0.234 & 1391 \\ \hline
        42 & K. De Bruyne & Manchester City & England. First & 2019/2020 & 0.234 & 2133 \\ \hline
        43 & Francisco Conceicão & Porto & Portugal. First & 2023/2024 & 0.233 & 1133 \\ \hline
        44 & V. Birmančević & Cukaricki & Serbia. First & 2020/2021 & 0.233 & 1217 \\ \hline
        45 & A. Gómez & Atalanta & Italy. First & 2019/2020 & 0.233 & 2095 \\ \hline
        46 & Neymar & PSG & France. First & 2020/2021 & 0.233 & 1051 \\ \hline
        47 & M. Melikson & Hapoel Be er Sheva & Israel. First & 2016/2017 & 0.232 & 1341 \\ \hline
        48 & Cesc Fàbregas & Chelsea & England. First & 2016/2017 & 0.231 & 1006 \\ \hline
        49 & M. Tomasov & Astana & Kazakhstan. First & 2023 & 0.231 & 1347 \\ \hline
        50 & N. Lang & Club Brugge & Belgium. First & 2021/2022 & 0.231 & 2052 \\ \hline
    \end{tabular}
    \label{historical_players_epv}
\end{table}

\begin{table}[H]
    \centering
    \caption{Player Ranking Based On PCR Predictions In The Case Of Their Transfer To Manchester City}
        \begin{tabular}{|l|l|l|l|l|l|l|l|}
    \hline
        \# & player & team & age & PCR & PCR\_pred & PCR\_adj & stay\_proba \\ \hline
        1 & J. Doku & Manchester City & 21.2 & 0.299 & 0.197 & 0.196 & 0.961 \\ \hline
        2 & O. Dembélé & PSG & 26.3 & 0.215 & 0.173 & 0.166 & 0.868 \\ \hline
        3 & Sávio & Girona & 19.4 & 0.174 & 0.165 & 0.162 & 0.96 \\ \hline
        4 & Nico Williams & Athletic Bilbao & 21.1 & 0.199 & 0.157 & 0.154 & 0.956 \\ \hline
        5 & Bryan Zaragoza & Bayern Munchen & 21.9 & 0.162 & 0.154 & 0.151 & 0.961 \\ \hline
        6 & K. Kvaratskhelia & Napoli & 22.5 & 0.195 & 0.153 & 0.15 & 0.954 \\ \hline
        7 & Francisco Conceicão & Porto & 20.7 & 0.233 & 0.156 & 0.149 & 0.905 \\ \hline
        8 & M. Olise & Crystal Palace & 21.7 & 0.182 & 0.152 & 0.148 & 0.862 \\ \hline
        9 & O. Sahraoui & Heerenveen & 22.2 & 0.151 & 0.155 & 0.146 & 0.882 \\ \hline
        10 & Vinícius Júnior & Real Madrid & 23.1 & 0.136 & 0.148 & 0.144 & 0.938 \\ \hline
        11 & K. Coman & Bayern Munchen & 27.2 & 0.095 & 0.149 & 0.144 & 0.861 \\ \hline
        12 & R. Sterling & Chelsea & 28.7 & 0.152 & 0.149 & 0.144 & 0.828 \\ \hline
        13 & O. Niang & Riga & 21.4 & 0.296 & 0.161 & 0.144 & 0.734 \\ \hline
        14 & C. Ejuke & Antwerp & 25.6 & 0.158 & 0.152 & 0.143 & 0.848 \\ \hline
        15 & N. Lang & PSV & 24.2 & 0.196 & 0.156 & 0.143 & 0.74 \\ \hline
        16 & Rafael Leão & Milan & 24.2 & 0.178 & 0.145 & 0.142 & 0.913 \\ \hline
        17 & J. Dompé & Hamburger SV & 28.0 & 0.158 & 0.155 & 0.141 & 0.768 \\ \hline
        18 & F. Chiesa & Juventus & 25.8 & 0.195 & 0.144 & 0.141 & 0.932 \\ \hline
        19 & M. Edwards & Sporting CP & 24.7 & 0.277 & 0.149 & 0.141 & 0.866 \\ \hline
        20 & T. Corbeanu & Granada & 21.2 & 0.152 & 0.148 & 0.14 & 0.869 \\ \hline
        21 & B. Gruda & Mainz 05 & 19.2 & 0.163 & 0.144 & 0.138 & 0.823 \\ \hline
        22 & S. Ltaief & Winterthur & 23.3 & 0.114 & 0.148 & 0.137 & 0.786 \\ \hline
        23 & I. Akhomach & Villarreal & 19.3 & 0.094 & 0.14 & 0.137 & 0.936 \\ \hline
        24 & J. Enciso & Brighton & 19.6 & 0.215 & 0.147 & 0.137 & 0.677 \\ \hline
        25 & Pedro Neto & Wolverhampton & 23.4 & 0.131 & 0.14 & 0.136 & 0.855 \\ \hline
        26 & T. Kubo & Real Sociedad & 22.2 & 0.162 & 0.139 & 0.135 & 0.933 \\ \hline
        27 & T. Ali & Malmo FF & 24.8 & 0.277 & 0.145 & 0.135 & 0.828 \\ \hline
        28 & J. Bynoe-Gittens & Borussia Dortmund & 19.0 & 0.208 & 0.139 & 0.135 & 0.898 \\ \hline
        29 & J. Sancho & Borussia Dortmund & 23.4 & 0.224 & 0.14 & 0.135 & 0.858 \\ \hline
        30 & A. Mitriță & Universitatea Craiova & 28.4 & 0.195 & 0.143 & 0.135 & 0.906 \\ \hline
        31 & A. Nusa & Club Brugge & 18.3 & 0.143 & 0.142 & 0.134 & 0.872 \\ \hline
        32 & J. Bakayoko & PSV & 20.3 & 0.194 & 0.141 & 0.134 & 0.928 \\ \hline
        33 & E. Zhegrova & Lille & 24.4 & 0.097 & 0.139 & 0.134 & 0.907 \\ \hline
        34 & Y. Minteh & Feyenoord & 19.1 & 0.217 & 0.145 & 0.134 & 0.79 \\ \hline
        35 & Lamine Yamal & Barcelona & 16.1 & 0.136 & 0.138 & 0.134 & 0.922 \\ \hline
        36 & Haissem Hassan & Sporting Gijon & 21.5 & 0.125 & 0.144 & 0.133 & 0.873 \\ \hline
        37 & K. Mitoma & Brighton & 26.2 & 0.215 & 0.137 & 0.133 & 0.856 \\ \hline
        38 & A. Sbaï & Grenoble & 22.8 & 0.185 & 0.145 & 0.132 & 0.818 \\ \hline
        39 & Serginho & Viborg & 22.5 & 0.182 & 0.144 & 0.132 & 0.719 \\ \hline
        40 & R. Sottil & Fiorentina & 24.2 & 0.097 & 0.137 & 0.132 & 0.84 \\ \hline
        41 & A. Schjelderup & Nordsjaelland & 19.2 & 0.283 & 0.137 & 0.131 & 0.936 \\ \hline
        42 & N. Madueke & Chelsea & 21.4 & 0.177 & 0.136 & 0.131 & 0.823 \\ \hline
        43 & M. Fofana & Lyon & 18.3 & 0.169 & 0.136 & 0.131 & 0.93 \\ \hline
        44 & Milson & Maccabi Tel Aviv & 23.9 & 0.178 & 0.14 & 0.13 & 0.856 \\ \hline
        45 & Ângelo Gabriel & Strasbourg & 18.6 & 0.125 & 0.134 & 0.13 & 0.918 \\ \hline
        46 & Dudu & Palmeiras & 31.3 & 0.172 & 0.147 & 0.13 & 0.57 \\ \hline
        47 & Alejandro Garnacho & Manchester United & 19.1 & 0.088 & 0.132 & 0.129 & 0.922 \\ \hline
        48 & David Neres & Benfica Portugal & 26.5 & 0.25 & 0.142 & 0.129 & 0.708 \\ \hline
        49 & Gabriel Martinelli & Arsenal & 22.2 & 0.132 & 0.133 & 0.129 & 0.874 \\ \hline
        50 & Luis Guilherme & Palmeiras & 17.2 & 0.127 & 0.144 & 0.129 & 0.62 \\ \hline
    \end{tabular}
    \label{PCR_predictionsManchesterCity}
\end{table}

\begin{table}[H]
    \centering
    \caption{Player Ranking Based On PCR Predictions In The Case Of Their Transfer To Barcelona}
        \begin{tabular}{|l|l|l|l|l|l|l|l|}
    \hline
        \# & player & team & age & PCR & PCR\_pred & PCR\_adj & stay\_proba \\ \hline
        1 & J. Doku & Manchester City & 21.2 & 0.299 & 0.182 & 0.182 & 0.961 \\ \hline
        2 & O. Dembélé & PSG & 26.3 & 0.215 & 0.179 & 0.173 & 0.868 \\ \hline
        3 & Sávio & Girona & 19.4 & 0.174 & 0.163 & 0.162 & 0.96 \\ \hline
        4 & Nico Williams & Athletic Bilbao & 21.1 & 0.199 & 0.155 & 0.153 & 0.956 \\ \hline
        5 & Bryan Zaragoza & Bayern Munchen & 21.9 & 0.162 & 0.153 & 0.152 & 0.961 \\ \hline
        6 & K. Kvaratskhelia & Napoli & 22.5 & 0.195 & 0.151 & 0.15 & 0.954 \\ \hline
        7 & Vinícius Júnior & Real Madrid & 23.1 & 0.136 & 0.148 & 0.146 & 0.938 \\ \hline
        8 & E. Zhegrova & Lille & 24.4 & 0.097 & 0.147 & 0.143 & 0.907 \\ \hline
        9 & M. Olise & Crystal Palace & 21.7 & 0.182 & 0.146 & 0.143 & 0.862 \\ \hline
        10 & Lamine Yamal & Barcelona & 16.1 & 0.136 & 0.144 & 0.141 & 0.922 \\ \hline
        11 & N. Lang & PSV & 24.2 & 0.196 & 0.152 & 0.14 & 0.74 \\ \hline
        12 & T. Kubo & Real Sociedad & 22.2 & 0.162 & 0.142 & 0.14 & 0.933 \\ \hline
        13 & I. Akhomach & Villarreal & 19.3 & 0.094 & 0.142 & 0.14 & 0.936 \\ \hline
        14 & Rafael Leão & Milan & 24.2 & 0.178 & 0.142 & 0.14 & 0.913 \\ \hline
        15 & C. Ejuke & Antwerp & 25.6 & 0.158 & 0.146 & 0.14 & 0.848 \\ \hline
        16 & R. Sterling & Chelsea & 28.7 & 0.152 & 0.143 & 0.139 & 0.828 \\ \hline
        17 & Francisco Conceicão & Porto & 20.7 & 0.233 & 0.144 & 0.139 & 0.905 \\ \hline
        18 & O. Niang & Riga & 21.4 & 0.296 & 0.153 & 0.138 & 0.734 \\ \hline
        19 & O. Sahraoui & Heerenveen & 22.2 & 0.151 & 0.145 & 0.138 & 0.882 \\ \hline
        20 & F. Chiesa & Juventus & 25.8 & 0.195 & 0.139 & 0.137 & 0.932 \\ \hline
        21 & K. Coman & Bayern Munchen & 27.2 & 0.095 & 0.14 & 0.137 & 0.861 \\ \hline
        22 & T. Corbeanu & Granada & 21.2 & 0.152 & 0.143 & 0.137 & 0.869 \\ \hline
        23 & J. Dompé & Hamburger SV & 28.0 & 0.158 & 0.147 & 0.136 & 0.768 \\ \hline
        24 & Pedro Neto & Wolverhampton & 23.4 & 0.131 & 0.138 & 0.135 & 0.855 \\ \hline
        25 & J. Sancho & Borussia Dortmund & 23.4 & 0.224 & 0.138 & 0.135 & 0.858 \\ \hline
        26 & T. Ali & Malmo FF & 24.8 & 0.277 & 0.142 & 0.134 & 0.828 \\ \hline
        27 & M. Edwards & Sporting CP & 24.7 & 0.277 & 0.14 & 0.133 & 0.866 \\ \hline
        28 & Haissem Hassan & Sporting Gijon & 21.5 & 0.125 & 0.141 & 0.133 & 0.873 \\ \hline
        29 & Ângelo Gabriel & Strasbourg & 18.6 & 0.125 & 0.135 & 0.132 & 0.918 \\ \hline
        30 & J. Bynoe-Gittens & Borussia Dortmund & 19.0 & 0.208 & 0.134 & 0.132 & 0.898 \\ \hline
        31 & R. Sottil & Fiorentina & 24.2 & 0.097 & 0.135 & 0.13 & 0.84 \\ \hline
        32 & K. Mitoma & Brighton & 26.2 & 0.215 & 0.133 & 0.13 & 0.856 \\ \hline
        33 & A. Nusa & Club Brugge & 18.3 & 0.143 & 0.135 & 0.129 & 0.872 \\ \hline
        34 & M. Politano & Napoli & 30.1 & 0.188 & 0.134 & 0.129 & 0.796 \\ \hline
        35 & B. Gruda & Mainz 05 & 19.2 & 0.163 & 0.133 & 0.128 & 0.823 \\ \hline
        36 & J. Bakayoko & PSV & 20.3 & 0.194 & 0.133 & 0.128 & 0.928 \\ \hline
        37 & Ez Abde & Real Betis & 21.7 & 0.126 & 0.13 & 0.128 & 0.948 \\ \hline
        38 & N. Madueke & Chelsea & 21.4 & 0.177 & 0.132 & 0.128 & 0.823 \\ \hline
        39 & Alejandro Garnacho & Manchester United & 19.1 & 0.088 & 0.129 & 0.128 & 0.922 \\ \hline
        40 & M. Fofana & Lyon & 18.3 & 0.169 & 0.131 & 0.127 & 0.93 \\ \hline
        41 & A. Mitriță & Universitatea Craiova & 28.4 & 0.195 & 0.133 & 0.127 & 0.906 \\ \hline
        42 & A. Schjelderup & Nordsjaelland & 19.2 & 0.283 & 0.13 & 0.127 & 0.936 \\ \hline
        43 & Milson & Maccabi Tel Aviv & 23.9 & 0.178 & 0.134 & 0.126 & 0.856 \\ \hline
        44 & S. Ltaief & Winterthur & 23.3 & 0.114 & 0.134 & 0.126 & 0.786 \\ \hline
        45 & M. Simon & Nantes & 28.1 & 0.125 & 0.132 & 0.126 & 0.801 \\ \hline
        46 & J. Boga & Nice & 26.6 & 0.079 & 0.129 & 0.125 & 0.894 \\ \hline
        47 & Dudu & Palmeiras & 31.3 & 0.172 & 0.14 & 0.125 & 0.57 \\ \hline
        48 & J. Clarke & Sunderland & 22.7 & 0.268 & 0.131 & 0.125 & 0.939 \\ \hline
        49 & Serginho & Viborg & 22.5 & 0.182 & 0.135 & 0.124 & 0.719 \\ \hline
        50 & M. Daramy & Reims & 21.6 & 0.077 & 0.127 & 0.124 & 0.926 \\ \hline
    \end{tabular}

    \label{PCR_predictionsBarcelona}
\end{table}

\begin{table}[H]
    \centering
    \caption{Player Ranking Based On PCR Predictions In The Case Of Their Transfer To Milan}
    \begin{tabular}{|l|l|l|l|l|l|l|l|}
\hline
    \# & player & team & age & PCR & PCR\_pred & PCR\_adj & stay\_proba \\ \hline
    1 & J. Doku & Manchester City & 21.2 & 0.299 & 0.173 & 0.173 & 0.961 \\ \hline
    2 & O. Dembélé & PSG & 26.3 & 0.215 & 0.168 & 0.162 & 0.868 \\ \hline
    3 & Sávio & Girona & 19.4 & 0.174 & 0.159 & 0.157 & 0.96 \\ \hline
    4 & Nico Williams & Athletic Bilbao & 21.1 & 0.199 & 0.151 & 0.15 & 0.956 \\ \hline
    5 & K. Kvaratskhelia & Napoli & 22.5 & 0.195 & 0.148 & 0.147 & 0.954 \\ \hline
    6 & Bryan Zaragoza & Bayern Munchen & 21.9 & 0.162 & 0.147 & 0.145 & 0.961 \\ \hline
    7 & Vinícius Júnior & Real Madrid & 23.1 & 0.136 & 0.145 & 0.142 & 0.938 \\ \hline
    8 & Rafael Leão & Milan & 24.2 & 0.178 & 0.145 & 0.142 & 0.913 \\ \hline
    9 & R. Sterling & Chelsea & 28.7 & 0.152 & 0.142 & 0.137 & 0.828 \\ \hline
    10 & M. Olise & Crystal Palace & 21.7 & 0.182 & 0.141 & 0.137 & 0.862 \\ \hline
    11 & N. Lang & PSV & 24.2 & 0.196 & 0.149 & 0.137 & 0.74 \\ \hline
    12 & O. Sahraoui & Heerenveen & 22.2 & 0.151 & 0.144 & 0.137 & 0.882 \\ \hline
    13 & C. Ejuke & Antwerp & 25.6 & 0.158 & 0.144 & 0.137 & 0.848 \\ \hline
    14 & T. Ali & Malmo FF & 24.8 & 0.277 & 0.143 & 0.134 & 0.828 \\ \hline
    15 & K. Coman & Bayern Munchen & 27.2 & 0.095 & 0.137 & 0.133 & 0.861 \\ \hline
    16 & J. Dompé & Hamburger SV & 28.0 & 0.158 & 0.145 & 0.133 & 0.768 \\ \hline
    17 & T. Kubo & Real Sociedad & 22.2 & 0.162 & 0.135 & 0.133 & 0.933 \\ \hline
    18 & T. Corbeanu & Granada & 21.2 & 0.152 & 0.139 & 0.132 & 0.869 \\ \hline
    19 & Francisco Conceicão & Porto & 20.7 & 0.233 & 0.137 & 0.132 & 0.905 \\ \hline
    20 & Pedro Neto & Wolverhampton & 23.4 & 0.131 & 0.135 & 0.132 & 0.855 \\ \hline
    21 & E. Zhegrova & Lille & 24.4 & 0.097 & 0.135 & 0.131 & 0.907 \\ \hline
    22 & F. Chiesa & Juventus & 25.8 & 0.195 & 0.133 & 0.131 & 0.932 \\ \hline
    23 & K. Mitoma & Brighton & 26.2 & 0.215 & 0.134 & 0.131 & 0.856 \\ \hline
    24 & I. Akhomach & Villarreal & 19.3 & 0.094 & 0.133 & 0.131 & 0.936 \\ \hline
    25 & J. Sancho & Borussia Dortmund & 23.4 & 0.224 & 0.134 & 0.13 & 0.858 \\ \hline
    26 & Haissem Hassan & Sporting Gijon & 21.5 & 0.125 & 0.139 & 0.13 & 0.873 \\ \hline
    27 & O. Niang & Riga & 21.4 & 0.296 & 0.144 & 0.129 & 0.734 \\ \hline
    28 & J. Bynoe-Gittens & Borussia Dortmund & 19.0 & 0.208 & 0.132 & 0.129 & 0.898 \\ \hline
    29 & R. Sottil & Fiorentina & 24.2 & 0.097 & 0.134 & 0.129 & 0.84 \\ \hline
    30 & M. Edwards & Sporting CP & 24.7 & 0.277 & 0.134 & 0.128 & 0.866 \\ \hline
    31 & Lamine Yamal & Barcelona & 16.1 & 0.136 & 0.129 & 0.127 & 0.922 \\ \hline
    32 & Ângelo Gabriel & Strasbourg & 18.6 & 0.125 & 0.129 & 0.126 & 0.918 \\ \hline
    33 & M. Politano & Napoli & 30.1 & 0.188 & 0.132 & 0.126 & 0.796 \\ \hline
    34 & Milson & Maccabi Tel Aviv & 23.9 & 0.178 & 0.134 & 0.126 & 0.856 \\ \hline
    35 & Ez Abde & Real Betis & 21.7 & 0.126 & 0.127 & 0.125 & 0.948 \\ \hline
    36 & B. Gruda & Mainz 05 & 19.2 & 0.163 & 0.128 & 0.124 & 0.823 \\ \hline
    37 & Alejandro Garnacho & Manchester United & 19.1 & 0.088 & 0.125 & 0.124 & 0.922 \\ \hline
    38 & A. Nusa & Club Brugge & 18.3 & 0.143 & 0.129 & 0.124 & 0.872 \\ \hline
    39 & N. Madueke & Chelsea & 21.4 & 0.177 & 0.127 & 0.123 & 0.823 \\ \hline
    40 & J. Bakayoko & PSV & 20.3 & 0.194 & 0.128 & 0.123 & 0.928 \\ \hline
    41 & David Neres & Benfica Portugal & 26.5 & 0.25 & 0.133 & 0.122 & 0.708 \\ \hline
    42 & J. Clarke & Sunderland & 22.7 & 0.268 & 0.128 & 0.122 & 0.939 \\ \hline
    43 & A. Mitriță & Universitatea Craiova & 28.4 & 0.195 & 0.128 & 0.122 & 0.906 \\ \hline
    44 & Daniel Podence & Olympiacos Piraeus & 27.8 & 0.252 & 0.131 & 0.121 & 0.8 \\ \hline
    45 & A. Schjelderup & Nordsjaelland & 19.2 & 0.283 & 0.125 & 0.121 & 0.936 \\ \hline
    46 & Dudu & Palmeiras & 31.3 & 0.172 & 0.136 & 0.121 & 0.57 \\ \hline
    47 & M. Fofana & Lyon & 18.3 & 0.169 & 0.125 & 0.121 & 0.93 \\ \hline
    48 & Serginho & Viborg & 22.5 & 0.182 & 0.129 & 0.119 & 0.719 \\ \hline
    49 & A. Sbaï & Grenoble & 22.8 & 0.185 & 0.128 & 0.118 & 0.818 \\ \hline
    50 & J. Boga & Nice & 26.6 & 0.079 & 0.122 & 0.118 & 0.894 \\ \hline
\end{tabular}

    \label{PCR_predictionsMilan}
\end{table}

\begin{table}[H]
    \centering
    \caption{Player Ranking Based On PCR Predictions In The Case Of Their Transfer To Brighton}
        \begin{tabular}{|l|l|l|l|l|l|l|l|}
    \hline
        \# & player & team & age & PCR & PCR\_pred & PCR\_adj & stay\_proba \\ \hline
        1 & J. Doku & Manchester City & 21.2 & 0.299 & 0.161 & 0.159 & 0.961 \\ \hline
        2 & O. Dembélé & PSG & 26.3 & 0.215 & 0.15 & 0.143 & 0.868 \\ \hline
        3 & Sávio & Girona & 19.4 & 0.174 & 0.141 & 0.139 & 0.96 \\ \hline
        4 & R. Sterling & Chelsea & 28.7 & 0.152 & 0.134 & 0.129 & 0.828 \\ \hline
        5 & K. Kvaratskhelia & Napoli & 22.5 & 0.195 & 0.13 & 0.127 & 0.954 \\ \hline
        6 & Nico Williams & Athletic Bilbao & 21.1 & 0.199 & 0.129 & 0.127 & 0.956 \\ \hline
        7 & M. Olise & Crystal Palace & 21.7 & 0.182 & 0.13 & 0.126 & 0.862 \\ \hline
        8 & J. Dompé & Hamburger SV & 28.0 & 0.158 & 0.138 & 0.126 & 0.768 \\ \hline
        9 & C. Ejuke & Antwerp & 25.6 & 0.158 & 0.133 & 0.125 & 0.848 \\ \hline
        10 & O. Sahraoui & Heerenveen & 22.2 & 0.151 & 0.131 & 0.124 & 0.882 \\ \hline
        11 & Bryan Zaragoza & Bayern Munchen & 21.9 & 0.162 & 0.126 & 0.124 & 0.961 \\ \hline
        12 & Haissem Hassan & Sporting Gijon & 21.5 & 0.125 & 0.132 & 0.123 & 0.873 \\ \hline
        13 & T. Corbeanu & Granada & 21.2 & 0.152 & 0.127 & 0.12 & 0.869 \\ \hline
        14 & Vinícius Júnior & Real Madrid & 23.1 & 0.136 & 0.123 & 0.12 & 0.938 \\ \hline
        15 & T. Ali & Malmo FF & 24.8 & 0.277 & 0.127 & 0.119 & 0.828 \\ \hline
        16 & N. Lang & PSV & 24.2 & 0.196 & 0.128 & 0.117 & 0.74 \\ \hline
        17 & K. Mitoma & Brighton & 26.2 & 0.215 & 0.12 & 0.117 & 0.856 \\ \hline
        18 & O. Niang & Riga & 21.4 & 0.296 & 0.13 & 0.116 & 0.734 \\ \hline
        19 & R. Sottil & Fiorentina & 24.2 & 0.097 & 0.121 & 0.116 & 0.84 \\ \hline
        20 & Pedro Neto & Wolverhampton & 23.4 & 0.131 & 0.12 & 0.116 & 0.855 \\ \hline
        21 & I. Akhomach & Villarreal & 19.3 & 0.094 & 0.119 & 0.116 & 0.936 \\ \hline
        22 & Rafael Leão & Milan & 24.2 & 0.178 & 0.118 & 0.115 & 0.913 \\ \hline
        23 & Francisco Conceicão & Porto & 20.7 & 0.233 & 0.12 & 0.114 & 0.905 \\ \hline
        24 & K. Coman & Bayern Munchen & 27.2 & 0.095 & 0.118 & 0.114 & 0.861 \\ \hline
        25 & Ângelo Gabriel & Strasbourg & 18.6 & 0.125 & 0.117 & 0.114 & 0.918 \\ \hline
        26 & E. Zhegrova & Lille & 24.4 & 0.097 & 0.117 & 0.113 & 0.907 \\ \hline
        27 & A. Mitriță & Universitatea Craiova & 28.4 & 0.195 & 0.12 & 0.113 & 0.906 \\ \hline
        28 & J. Clarke & Sunderland & 22.7 & 0.268 & 0.119 & 0.112 & 0.939 \\ \hline
        29 & S. Ltaief & Winterthur & 23.3 & 0.114 & 0.121 & 0.112 & 0.786 \\ \hline
        30 & A. Nusa & Club Brugge & 18.3 & 0.143 & 0.118 & 0.112 & 0.872 \\ \hline
        31 & N. Madueke & Chelsea & 21.4 & 0.177 & 0.116 & 0.112 & 0.823 \\ \hline
        32 & Milson & Maccabi Tel Aviv & 23.9 & 0.178 & 0.12 & 0.112 & 0.856 \\ \hline
        33 & M. Edwards & Sporting CP & 24.7 & 0.277 & 0.118 & 0.111 & 0.866 \\ \hline
        34 & F. Chiesa & Juventus & 25.8 & 0.195 & 0.114 & 0.111 & 0.932 \\ \hline
        35 & J. Boga & Nice & 26.6 & 0.079 & 0.115 & 0.111 & 0.894 \\ \hline
        36 & Alejandro Garnacho & Manchester United & 19.1 & 0.088 & 0.112 & 0.111 & 0.922 \\ \hline
        37 & T. Kubo & Real Sociedad & 22.2 & 0.162 & 0.113 & 0.11 & 0.933 \\ \hline
        38 & J. Bynoe-Gittens & Borussia Dortmund & 19.0 & 0.208 & 0.112 & 0.109 & 0.898 \\ \hline
        39 & A. Schjelderup & Nordsjaelland & 19.2 & 0.283 & 0.113 & 0.109 & 0.936 \\ \hline
        40 & Serginho & Viborg & 22.5 & 0.182 & 0.119 & 0.109 & 0.719 \\ \hline
        41 & M. Fofana & Lyon & 18.3 & 0.169 & 0.113 & 0.108 & 0.93 \\ \hline
        42 & B. Gruda & Mainz 05 & 19.2 & 0.163 & 0.113 & 0.108 & 0.823 \\ \hline
        43 & Y. Soteldo & Santos & 25.8 & 0.27 & 0.125 & 0.108 & 0.485 \\ \hline
        44 & Dudu & Palmeiras & 31.3 & 0.172 & 0.122 & 0.108 & 0.57 \\ \hline
        45 & J. Sancho & Borussia Dortmund & 23.4 & 0.224 & 0.111 & 0.107 & 0.858 \\ \hline
        46 & Danilo Al Saed & Sandefjord & 24.1 & 0.11 & 0.118 & 0.107 & 0.741 \\ \hline
        47 & C. Hudson-Odoi & Nottingham Forest & 22.8 & 0.114 & 0.11 & 0.107 & 0.884 \\ \hline
        48 & J. Enciso & Brighton & 19.6 & 0.215 & 0.114 & 0.106 & 0.677 \\ \hline
        49 & A. Sbaï & Grenoble & 22.8 & 0.185 & 0.116 & 0.106 & 0.818 \\ \hline
        50 & David Neres & Benfica Portugal & 26.5 & 0.25 & 0.117 & 0.106 & 0.708 \\ \hline
    \end{tabular}
    \label{PCR_predictionsBrighton}
\end{table}

\end{document}